\documentclass[numsec,webpdf,contemporary,large]{oup-authoring-template}

\usepackage[utf8]{inputenc}
\usepackage[T1]{fontenc}
\usepackage{textcomp}
\usepackage[english]{babel}
\usepackage{amsmath, amssymb,amsthm}
\usepackage{hyperref}
\usepackage{graphicx}
\usepackage{caption}
\usepackage{subcaption}
\usepackage{multirow}
\usepackage{threeparttable} 
\usepackage{footnote}
\makesavenoteenv{table}
\usepackage[linesnumbered,ruled,vlined]{algorithm2e}

\usepackage{import}
\usepackage{xifthen}
\usepackage{transparent}
\usepackage{tikz}
\usetikzlibrary{fit,positioning}
\graphicspath{ {./images/} }

\usepackage{bbm,accents} 
\usepackage{amsthm}

\newcommand{\mb}{\mathbf}

\newcommand{\mc}{\mathcal}

\newcommand{\bb}{\mathbb}

\newcommand{\set}[1]{\left\{ #1 \right\}}

\newcommand{\norm}[2]{\left\| #1 \right\|_{#2}}
\newcommand{\abs}[1]{\left| #1 \right|}

\renewcommand{\mathbf}{\boldsymbol}

\DeclareMathOperator{\diag}{diag}

\begin{document}
\journaltitle{GLBIO 2025}
\DOI{DOI}
\copyrightyear{2025}
\pubyear{2015}
\access{}
\appnotes{Paper}

\firstpage{1}


\title[Deep Active Learning for Host Targeted Therapeutics]{Deep Active Learning based Experimental Design to Uncover Synergistic Genetic Interactions for Host Targeted Therapeutics}

\author[1,$\ast$]{Haonan Zhu}
\author[1]{Mary Silva}
\author[1]{Jose Cadena} 
\author[1]{Braden Soper}
\author[2,3]{Michał Lisicki}
\author[4]{Braian Peetoom}
\author[4]{Sergio E. Baranzini} 
\author[1]{Shivshankar Sundaram}
\author[1]{Priyadip Ray}
\author[1]{Jeff Drocco}

\authormark{Haonan Zhu et al.}

\address[1]{\orgname{Lawrence Livermore National Laboratory}, \orgaddress{\street{7000 East Ave}, \postcode{94550}, \state{CA}, \country{USA}}}
\address[2]{ \orgname{University of Guelph}, \orgaddress{\street{50 Stone Rd E}, \postcode{N1G 2W1}, \state{ON}, \country{Canada}}}
\address[3]{  \orgname{Vector Institute}, \orgaddress{\street{108 College St W1140}, \postcode{M5G 0C6}, \state{ON}, \country{Canada}}}
\address[4]{  \orgname{University of California San Francisco}, \orgaddress{\street{675 Nelson Rising Lane}, \postcode{94158}, \state{CA}, \country{USA}}}

\corresp[$\ast$]{Corresponding author. \href{email:zhu18@llnl.gov}{zhu18@llnl.gov}}





\abstract{Recent technological advances have introduced new high-throughput methods for studying host-virus interactions, but testing synergistic interactions between host gene pairs during infection remains relatively slow and labor intensive. Identification of multiple gene knockdowns that effectively inhibit viral replication requires a search over the combinatorial space of all possible target gene pairs and is infeasible via brute-force experiments. Although active learning methods for sequential experimental design have shown promise, existing approaches have generally been restricted to single-gene knockdowns or small-scale double knockdown datasets. In this study, we present an integrated Deep Active Learning (DeepAL) framework that incorporates information from a biological knowledge graph (SPOKE, the Scalable Precision Medicine Open Knowledge Engine) to efficiently search the configuration space of a large dataset of all pairwise knockdowns of 356 human genes in HIV infection. Through graph representation learning, the framework is able to generate task-specific representations of genes while also balancing the exploration-exploitation trade-off to pinpoint highly effective double-knockdown pairs.  We additionally present an ensemble method for uncertainty quantification and an interpretation of the gene pairs selected by our algorithm via pathway analysis. To our knowledge, this is the first work to show promising results on double-gene knockdown experimental data of appreciable scale (356 by 356 matrix).}

\maketitle






\section{Introduction}
\label{sec: intro}

Understanding the various pathways essential for viral replication is vital for creating effective antiviral treatments. Due to factors such as replication and mutation rates, many viruses quickly develop resistance to drugs targeting a single site, making them ineffective over time. In these situations, combination therapies are necessary to inhibit viral replication sufficiently to prevent escape through mutations \cite{tang2012hiv}. Although host cellular targets are not subject to the same selective pressures as viral genome components, targeting host factors crucial for viral replication may still benefit from a combination strategy. Viruses have developed intricate interactions with host cells to support their life cycle, and studies have shown that dual knockdowns of host genes can sometimes synergistically inhibit viral growth \cite{gordon2020quantitative}. Moreover, targeting multiple host factors may expand the effectiveness of a therapeutic across different viral pathogens \cite{richman2016antiviral}. 

Recent technological advances have introduced new high-throughput methods for studying host-virus interactions \cite{puschnik2017crispr}, but testing synergistic interactions between gene pairs during viral infections remains relatively slow and labor-intensive. Identification of promising gene-gene knockdowns requires a search in the combinatorial space of all possible target gene pairs and is infeasible via brute-force experimentation \cite{norman2019exploring, replogle2022mapping}. A potential solution is sequential experimental design guided by data-driven models to balance the trade-off between successful identification of relevant gene pairs and experimental cost \cite{sverchkov2017review}. Various approaches, including active learning (e.g., bandits \cite{pacchianoneural}) and traditional experimental design \cite{lyle2023discobax},  have been previously studied in the context of single gene knockdowns. See \cite{mehrjou2021genedisco} for an overview of the methods along with benchmark datasets for single genetic interventions. More recently, there has been related work in representation learning to discover pairwise genetic interactions from embeddings of single gene knockdowns \cite{jain2024active,jain2024automated}, and design of perturbation screens on RNA-seq data by incorporating prior knowledge from multiple data sources \cite{huang2024sequential}. However, the goal of \cite{jain2024active,jain2024automated} is to discover gene pairs that violate the additive assumption (i.e., the effect of a double gene knockdown is the sum of two single gene knockdowns) in double-knockdown experiments, and require measurements from microscopy images while incorporating no existing knowledge graph information to assist in the discovery; To predict the effects of unseen gene knockdown, \cite{huang2024sequential} operate on RNA-seq data, which have the major drawback that we can only observe perturbations that survive long enough to be measured. There has also been work that utilizes a large language model to design experiments \cite{huang2024crispr, roohani2024biodiscoveryagent}. While the initial results are promising, the number of relevant gene pairs that can be identified by the bio-agent is strictly limited due to API constraints and the tendency of large language models to generate hallucinations (i.e., plausible but factually incorrect or nonsensical information) \cite{xu2024hallucination}. In summary, although many proposed methods discuss extensions for double-knockdown experimental design, the experimental results from the current literature are limited to single-knockdown data \cite{huang2024sequential}, small-scale double-knockdown data (e.g., a search space of $50$ by $50$ matrix in \cite{jain2024active}, only $160$ out of $100576$ possible gene pairs are queried in \cite{roohani2024biodiscoveryagent}). 


We present an integrated deep active learning (DeepAL) framework that incorporates existing knowledge graph information to efficiently search the configuration space while balancing the exploration-exploitation trade-off through the active learning loop that utilizes an ensemble method for uncertainty quantification \cite{abdar2021review,lakshminarayanan2017simple}. Specifically, we leverage the Scalable Precision Medicine Open Knowledge Engine (SPOKE) \cite{morris2023scalable}. By using graph representation learning \cite{schlichtkrull2018modeling}, our framework is able to provide interpretable results on the set of gene pairs recommended by the framework. Most notably, this work differs from the previous graph learning-based approach ITERPER \cite{huang2024sequential} in the following manners, 1)  our GNN model is built for heterogeneous knowledge graphs such as SPOKE that includes different edge types \footnote{Gene regulatory interactions, gene-protein encoding, protein-protein interactions, and gene-biological processes} associations whereas the primary graphs used in ITERPER include the gene co-expression graph (a homogeneous graph constructed through thresholding the Pearson correlations) and gene-ontology graph (a bipartite graph between genes and a pathway term constructed through thresholding computed Jaccard indices) and 2) our work leverages ensemble method for uncertainty quantification instead of fitting a Gaussian-noise model, which has been shown to better capture the model uncertainty of deep learning models \cite{lakshminarayanan2017simple}. Our main contribution is the development of a unified active learning framework that includes a heterogeneous graph-based representation model that utilizes the existing knowledge graph and an ensemble method for uncertainty quantification. To our knowledge, this is the first work to show promising results on double genes knockdown experimental data of appreciable scale (356 by 356 matrix) \cite{gordon2020quantitative}. 

The rest of the paper is organized as follows: section \ref{sec: notation and data} introduces the notation and data (SPOKE and HIV dataset) that we use to evaluate our methods, section \ref{sec: methods} presents the proposed ensemble DeepAL framework, section \ref{sec: experiments} provides our experimental results on the HIV dataset, and section \ref{sec: conclusion} summarizes the findings of the article and discusses future directions.

\section{Notations and Data}
\label{sec: notation and data}

\subsection{Notations}
\label{subsec: notations}
We use bold uppercase letters for matrices, bold lowercase letters for vectors, and lowercase letters for scalars. The Hadamard (element-wise) product of vectors $\mb a$ and $\mb b$ is denoted by $\mb a \circ \mb b$, and $\diag$ denotes the function mapping a vector to a diagonal matrix with the components of the vector as its diagonal entries. We denote the Huber loss function \cite{huber1992robust} by:

\begin{equation}
\label{eqn: huber loss}
\text{Huber}\left( a \right) = \begin{cases}
    \frac{1}{2} a^2 & \text{if } |a| \leq 1, \\
    |a| - \frac{1}{2} & \text{if } |a| > 1,
\end{cases} 
\end{equation}

The activation function $\text{Softplus}$ is defined as:

\begin{equation}
\label{eqn: softplus}
\text{Softplus}\left( a \right) = \log(1+\exp(a))
\end{equation}

We denote the undirected knowledge graph as $\mc G=\left( \mc V, \mc E, \mc R \right)$, where $\mc V$ is the set of vertices, $\mc E$ is the set of edges and $\mc R$ is the set of relationships. Each node $v_i \in \mc V$ is associated with a label, $label(v) \in L$ that includes labeled edges. The edge labeled $\left(v_i, r, v_j \right) \in  \mc E$, where $i=1, \ldots, \abs{V}$ indexes the nodes, and $r \in \mc R$. Given a node $v_{i}$, the set of neighbor indices of node $v_i$ in relation $r \in \mc R$ is denoted as $\mc N_{i}^{r}$. The learned embedding for a given node $v_i \in \mc V$ is denoted as $\mb x_{i} \in \bb R^{d}$ where $d$ is the pre-assigned embedding dimension. The set of target genes (for example, HIV related genes in our experiments) is denoted as $H:=\set{v_1, \ldots, v_p} \subseteq \mc V$. The pairs of genes of interest are denoted as $S:= H \times H \subseteq \mc E$, and the matrix of viral loads (target) is denoted as $\mb Y \in \bb R^{p \times p}$, where $y_{ij}$ corresponds to the value of the viral loads when we down-regulate the gene $v_i$ and the gene $v_j$ of $H$.  


\subsection{HIV Dataset}
\label{subsec: HIV dataset}

Data on the genetic interaction of HIV from \cite{gordon2020quantitative} focus on understanding the genetic interactions that influence HIV infection. It includes quantitative measurements of how different genetic perturbations, primarily through CRISPR-based methods, interact to affect viral loads of HIV. In total, the interaction matrix consists of 356 genes prioritized for their established or potential roles in HIV viral replication, as identified through prior studies on host-pathogen interactions and HIV-related protein interactions.

\section{Methods}
\label{sec: methods}

\begin{figure}[t]
    \centering
    \includegraphics[width=0.9\columnwidth]{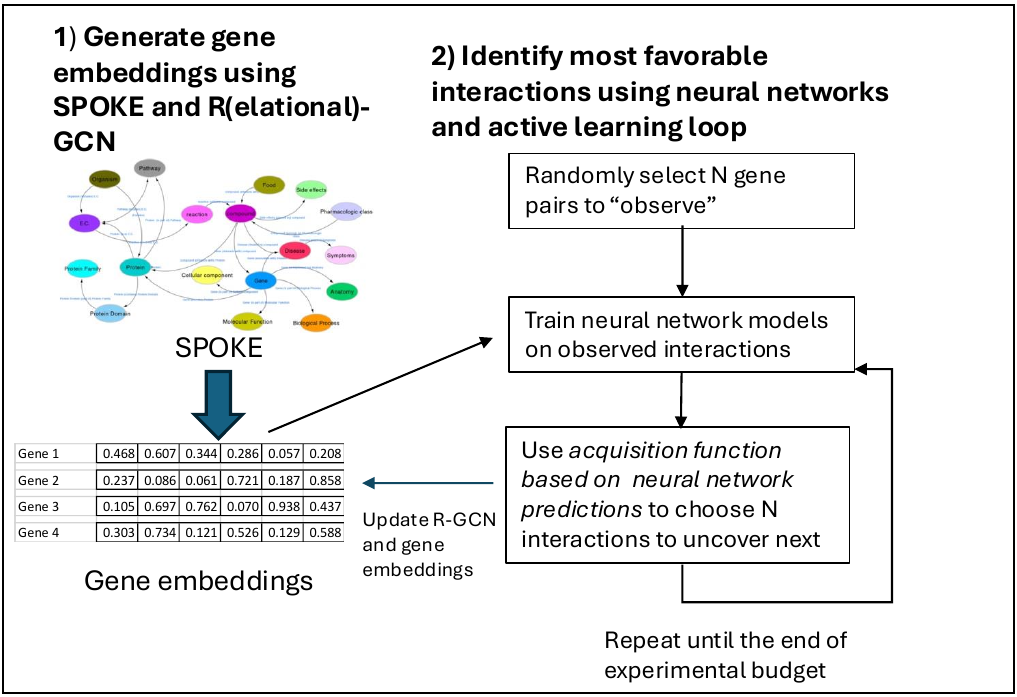}
    \caption{Flowchart of the proposed deep active learning framework. }
    \label{fig: deepAL flow chart}
\end{figure}

Our deep active learning framework incorporates information from the knowledge graph to efficiently explore the space of gene pairs. The framework has two phases: 1) initial self-supervised training of $M$-separate representation model(s) with different random initializations to learn the embeddings of genes that summarize the local topology of the knowledge graph,\footnote{$M=1$ for the base model}  2) an active learning loop where sequentially the representation model(s) and regression model(s) are further optimized to fit the current observed entries of the viral-loads matrix followed by the recommendations of the next batch of gene-pairs to uncover. The overall framework is summarized in Fig. \ref{fig: deepAL flow chart} and Algo. \ref{alg: DeepAL}. In Section \ref{subsec: model} we summarize the different components of the base model (the representation model and the regression model), in Section \ref{subsec: ensemble} we discuss uncertainty quantification via ensembles and in Section \ref{subsec: acquisition strategy} we discuss the acquisition strategy that is developed to balance the exploration-exploitation trade-off.

\begin{algorithm}[http]
\caption{DeepAL}
\label{alg: DeepAL}
\SetKwInOut{Require}{Require}
\SetKwFor{For}{for}{do}{end for}
\SetKwInput{KwData}{Require}
\SetKwInput{KwResult}{return}
\DontPrintSemicolon

\Require{knowledge graph \(\mc G=\left( \mc V, \mc E, \mc R \right)\), set of target genes $H$ with $p=\abs{H}$ and collection of gene-pairs of interests \(S \)}
\Require{viral-loads matrix \(\mb Y \in \bb R^{p \times p}\) to query}
\Require{number of rounds of experiments \(T\), number of gene-pairs to be selected per round \(N\)}
\Require{R-GCN models \(f_{\mb \Theta}^{(m)}: G \to \bb R^{\abs{V} \times d},\) where \(\mb \Theta\) denotes all the learnable parameters from the GNN model,  DistMulti weight matrix \(\mb R^{(m)} \in \bb R^{d \times d}\), bilinear regression weight matrices \(\mb A^{(m)} \in \bb R^{d \times d}\) and bias \(b^{(m)} \in \bb R\) , where \(m=1, \ldots, M\) denotes the distinct models with same architecture}
\Require{Acquisition Strategy, $\mb \Pi: S': \to S' \subseteq  S$ (see section \ref{subsec: acquisition strategy} for list of strategies considered)}

Train \(f_{\mb \Theta}^{(m)}\) and \(\mb R^{(m)} \) using self-supervised learning (see section \ref{subsec: model}) for each \(m=1, \ldots, M\)

Initialize $S'$ by randomly drawing \(N\) pairs of genes from \(S\)  

\For{\( t = 1, \ldots, T \) \tcp*{Active Learning Loop}}{

    \For{\(m=1, \ldots, M \) \tcp*{Train each model on \(S'\)}}{

    \( \min_{\mb \Theta^{(m)}, \mb A^{(m)}, b^{(m)}} \sum_{(i,j) \in S'} \text{Huber}(\hat{y}_{ij}^{(m)}-y_{ij})\), 
    }
    
    \(S_{new} \leftarrow \mb \Pi(S')\) 
    
    \( S' \leftarrow S' \cup S_{new} \)\;
}
\KwResult{observed set of gene-pairs \(S' \subseteq S \).}
\end{algorithm}

\subsection{SPOKE Knowledge Graph}
\label{subsec: spoke}
The SPOKE knowledge graph \cite{morris2023scalable} consists of more than 20 thousand human gene nodes encoding protein from Entrez Gene, more than 200 million human proteins, a combined 18 thousand nodes of biological processes, molecular function, and cellular components derived from the Gene Ontology database. Additionally, the graph includes over 1 million gene relationships in which the knockdown or knockdown of one gene, achieved by short hairpin RNA or CRISPR, results in the up-regulation or down-regulation of another gene, as indicated by consensus transcriptional profiles.

To focus on our modeling objectives, we derive a subgraph based on a random walk sampling procedure using the target gene nodes in our viral load matrix $\mb Y$ as the starting nodes in the random walk. This is effective for excluding nodes that are not closely related to the target genes, and similar ideas have been proposed to learn a meaningful latent representation of social networks \cite{perozzi2014deepwalk}. Given the subgraph $\mc G=\left( \mc V', \mc E', \mc R' \right)$ of interests from SPOKE, where $V'$ contains all the nodes belonging to genes, biological processes, proteins. For each target gene $v \in H$, perform $w=5$ independent random walks of length $s=5$, i.e., for each walk, we repeatedly select a node from neighbors of the current node with equal probability up to $s=5$ times. This results in a final set of nodes $\mc V$ that lies within the $5$-th order neighborhood of the target genes $H$ and a subgraph $\mc G=\left( \mc V, \mc E, \mc R \right)$ where all edges and relationships are restricted to the set of nodes $\mc V$.

\subsection{Base Model}
\label{subsec: model}

The proposed DeepAL framework contains three major components: 1) a representation model based on Relational Graph Convolutional Networks (R-GCNs) \cite{schlichtkrull2018modeling} that maps knowledge graph ($\mc G$) to node embeddings $\{\mb x_{i}\}_{v_{i}\in \mc V} $, 2) an edge prediction model based on DistMulti\cite{yang2014embedding} that maps a given pair of embeddings $\left( \mb x_{i},\mb x_{j} \right) $ to a binary prediction of whether the two nodes are related by relationship $r \in \mc R$, and 3) a bilinear regression model that maps a pair of node embeddings to the target variable of interest in active learning loop. The additional hyperparameter configurations are summarized in Table \ref{tab:hyperparameters}.  

\subsubsection{R-GCN}
\label{subsubsec: R-GCN}

The main propagation model for the forward-pass of a node (i.e., $v_{i}$) is:
\begin{equation}
\label{eqn: R-conv}
               \mb h_{i}^{\left(l+1\right)} = \sigma\left( \sum_{r \in \mc R} \sum_{j \in \mc N_{i}^{r}} \frac{1}{c_{i,r}} \mb W_{r}^{\left( l \right) } \mb h_{j}^{\left( l \right) } \right) + \mb W_{0}^{\left( l \right) }\mb h_{i}^{\left( l \right) } ,
\end{equation}
where $l=1,\ldots, L$ indexes the layer number and $L$ is the total number of convolution layers. $\mb h_i^l \in \bb R^{d_h}$ denotes embedding of node $i$ at the $l$'th layer, where $d_h$ is the dimension of the hidden layers, $c_{i,r}$ is a normalization constant (e.g., $c_{i,r}=\abs{N_{i}^{r}}$, the number of neighbors of node $v_{i}$). Each graph convolution operation aggregates feature vectors of each node's neighbors. In this work, we use the standard basis vectors for the initial embeddings (i.e., $\mb h_i^0=\mb e_i \in \bb R^{\abs{V}}$ for all $v_i \in \mc V$).

\subsubsection{DistMulti: Edge Prediction Model for Initialization} 
\label{subsubsec: DistMulti}

Every relation $r$ is associated with a diagonal matrix $\mb R_{r} \in \bb R^{d\times d}$, and for every pair of nodes $\left( v_{i}, v_{j} \right) $ is scored as:
\begin{equation}
\label{eqn: distmulti}
f\left( v_{i}, r, v_{j} \right)= \sigma\left( \mb x_{i}^{\top} \mb R_{r} \mb x_{j}  \right).
\end{equation}
The self-supervised learning task is performed based on negative sampling \cite{yang2014embedding, trouillon2016complex, schlichtkrull2018modeling} where for each observed relationship (that is, positive edge), we sample one false edge (i.e., negative edge), and both the representation model and DistMulti are jointly trained to classify these edges. 

\subsubsection{Bilinear Regression}
\label{subsubsec: bilinear}
For every pair of target nodes $\left( v_{i}, v_{j} \right) \in S$, the predicted  viral-loads are given by:
\begin{equation}
\label{eqn: bilinear reg}
\hat{y}_{ij}= \text{Softplus}\left(\mb x_{i}^{\top} \mb A \mb x_{j}\right)+b,
\end{equation}
where $\mb A$  is a trainable symmetric weight matrix and $b$ is trainable scalar for bias correction. This is a generalization of DistMulti, where the matrix is no longer restricted to be diagonal, and additional non-linearity is added to increase the expressiveness of the model. This has been studied extensively in the bilinear bandit literature \cite{jun2019bilinear,rizk2021best}. 

\begin{table}[h!]
\centering
\resizebox{0.95\columnwidth}{!}{
    \begin{tabular}{|c|c|c|c|c|}
    \hline
    \multirow{2}{*}{Model} & \multicolumn{4}{c|}{Hyperparameters} \\ \cline{2-5} 
                                & Normalization & Non-Linearlity $\sigma$ &  Regularization & Dimensional Parameters \\ \hline
    \multirow{3}{*}{R-GCN}     & \multirow{3}{*}{DiffGroupNorm \cite{zhou2020towards}}         & \multirow{3}{*}{ReLu}           & \multirow{3}{*}{Weight Decay}   & L=$3$ \footnote{This balances the trade-off between expressiveness of the model and the over-smoothing phenomenon where all nodes became indistinguishable in the embedding space \cite{rusch2023survey}.}          \\ \cline{5-5} 
                                &           &           &           & $d_h=64$         \\ \cline{5-5} 
                                &           &          &          & $d=50$        \\ \hline
    DistMulti    & N/A       & Sigmoid         & N/A  & N/A \\ \cline{2-5} \hline
    \multirow{2}{*}{Bilinear}     & \multirow{2}{*}{N/A}        &  \multirow{2}{*}{Softplus}        & Dropout \cite{srivastava2014dropout},        & \multirow{2}{*}{N/A}         \\
    &           &          &         Weight Decay &        \\ \hline
    \end{tabular}
    }
\caption{Summary of hyperparameter setups for different components of the framework.}
\label{tab:hyperparameters}
\end{table}


\subsection{Ensemble Method For Additional Uncertainty Quantification}
\label{subsec: ensemble}

Uncertainty is a critical part of active learning to balance the exploration-exploitation trade-off, and the ensemble method has been used with success for neural network-based models in supervised learning settings \cite{lakshminarayanan2017simple, abdar2021review}. 

Given the $M$-trained models, we have $M$-predictions ($\hat{Y}^{(m)}_{ij}, m=1, \ldots, M$, Eqn. \ref{eqn: bilinear reg}) for every pair of nodes $(v_i, v_j)$ that have not been uncovered. These predictions capture the model uncertainty. We can further compute estimators for standard deviations and quantiles.

\subsection{Acquisition Strategies}
\label{subsec: acquisition strategy}

We examine multiple acquisition strategies for both the base model and the ensemble model, and they are summarized in Table \ref{tab: acquisition strategies}. 

\begin{table*}[ht]
\scriptsize
\setlength{\tabcolsep}{4pt} 
\centering
\begin{tabular}{|p{3cm}|p{3.8cm}|p{5.3cm}|}
\hline 
\textbf{Acquisitions} & Base Model $M=1$ & Ensemble $M=20$ \\
\hline
Greedy &  $\arg\min_{(i,j) \in S'} \hat{y}_{ij}$ &  $\arg\min_{(i,j) \in S'} \text{median}_{m} \hat{y}_{ij}^{(m)}$ \\
\hline
Badge \footnotemark & $\arg\max_{(i,j) \in S'} \norm{\mb x_{i} \circ \mb x_{j}}{}$ &  $\arg\max_{(i,j) \in S'} \text{median}_m \norm{\mb x^{(m)}_{i} \circ \mb x_{j}}{}$\\
\hline
Optimism & N/A &  $\arg\min_{(i,j) \in S'} \text{quantile}_{m} \hat{y}_{ij}^{(m)}$\\
\hline
Maximum Variance &  N/A & $\arg\max_{(i,j) \in S'} \text{std}_{m} \hat{y}_{ij}^{(m)}$\\
\hline
\end{tabular}
\footnotetext{First proposed in \cite{ash2019deep} to use norm of the gradient from the last layer of the neural network for uncertainty quantification. This reduces to T(race)-optimality in optimal experimental design \cite{pukelsheim2006optimal} when there is no-hidden layers.} 
\caption{Comparison of acquisition strategies in the base model setting and ensemble setting. Both optimism and maximum variance requires uncertainty quantification hence unattainable in the base model setting.}
\label{tab: acquisition strategies}
\end{table*}



\section{Experiments}
\label{sec: experiments}

In this section, we evaluate the performance of our methods on the HIV dataset described in Section \ref{subsec: HIV dataset}, and our main result shows that by incorporating information from the knowledge graph and utilizing uncertainty quantification information through the ensemble approach, the proposed ensemble DeepAL framework uncovers 92\% of the top $400$ gene-pairs after observing less than $6.3\%$ of the entire matrix. 

\subsection{Performance Evaluation}
\label{subsec: performance evaluation}
To mimic real-world experimental needs, we run all algorithms for $17$ rounds (including the initial round of random selection), and at each round each algorithm uncovers $400$ gene pairs for observation, where the batch size of $400$  is chosen to approximately balance typical computational and experimental capabilities while allowing sufficient exploration of the interaction matrix. We evaluated the algorithms by two different evaluation metrics: 1) Coverage@400 defined as the fraction of top-$400$ gene pairs uncovered by each round; 2) MAE (Mean of Absolute Error), defined as the sum of absolute errors between the predicted values on the unseen data given the current model and the ground truth values at each round, divided by the total number of samples. The former evaluates how well the active learning framework can uncover the most relevant gene pairs, while the latter evaluates how well a predictive model was learned along the sequential process. 

\subsection{Comparison Among the Acquisition Strategies} 
\label{subsec: acquisition comparison}
\begin{figure*}[t]
    \centering
    \begin{subfigure}{0.9\columnwidth}\hspace*{-1em}
        \centering
		\includegraphics[width=\textwidth]{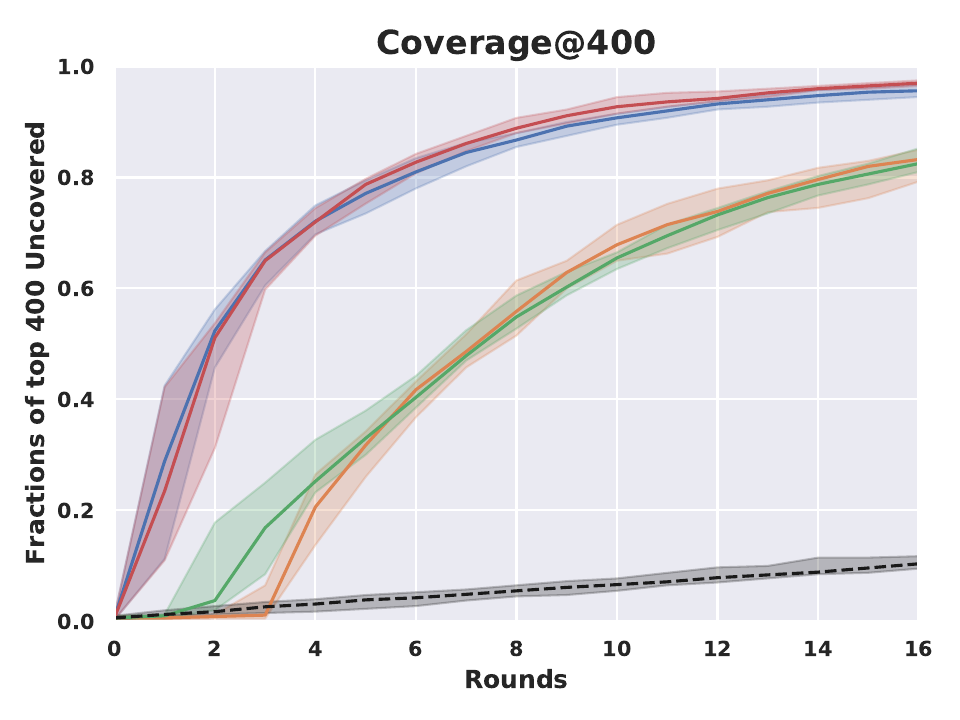}
		\caption{Coverage@400} 
    \end{subfigure}
    \begin{subfigure}{0.9\columnwidth}
        \centering
		\includegraphics[width=\textwidth]{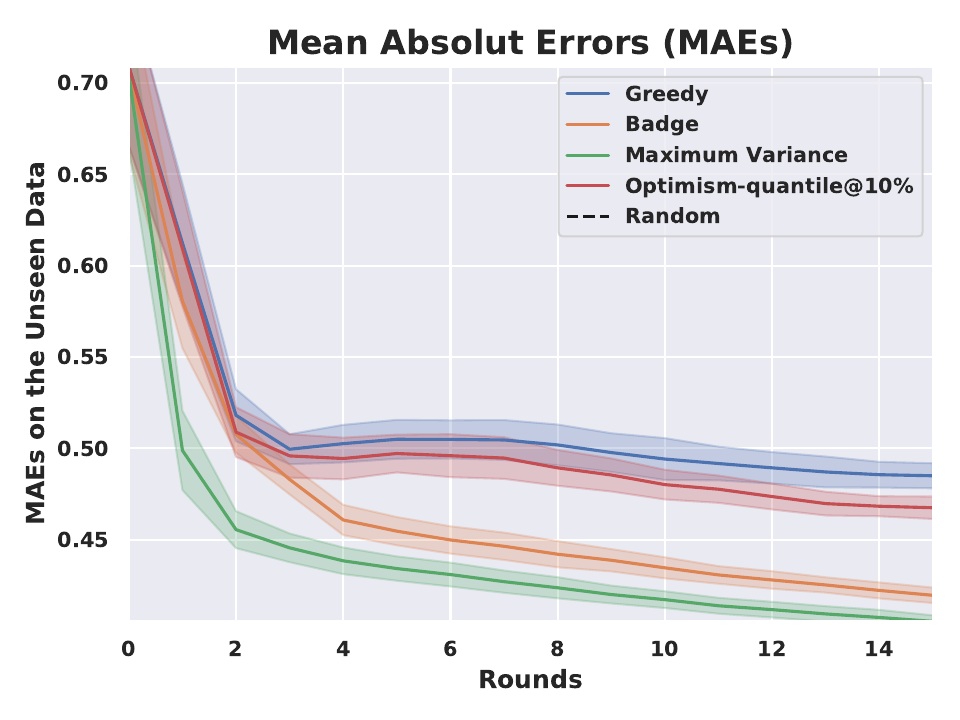}
		\caption{Mean Absolute Errors on Unseen Data}
    \end{subfigure}  
    \caption{Comparison among the different acquisition strategies, and results are summarized over $20$ replicates. Optimism with $10\%$ quantiles achieves the best performance in terms of coverage at terminal phase, and uncovers 92\% of the top $400$ gene-pairs while only $<6.3\%$ of the entire matrix is observed. Maximum variance strategy performs the best in learning a generalizable model that predicts viral-replicates on unseen data due to an emphasis on exploration.}
    \label{fig: acquisition comparisons}
\end{figure*}

We evaluate the list of acquisition functions detailed in Table. \ref{tab: acquisition strategies}, and the results are summarized in Fig. \ref{fig: acquisition comparisons}. Optimism with  $10\%$ quantiles achieves the best performance in terms of coverage at the end of the active sensing process, while the greedy solution offers competitive performance, especially in the early rounds. Badge and maximum variance strategies are suboptimal for coverage, but superior in learning a generalizable model that can predict viral-replicates on unseen data due to an emphasis on sampling gene-pairs where the predictive uncertainty is large (thus encouraging information gain over coverage). 

\subsection{Ablation Studies}
\label{subsec: abaltion studies}
\begin{figure*}[htbp]
    \centering
    \begin{subfigure}{0.9\columnwidth}\hspace*{-1em}
        \centering
		\includegraphics[width=\textwidth]{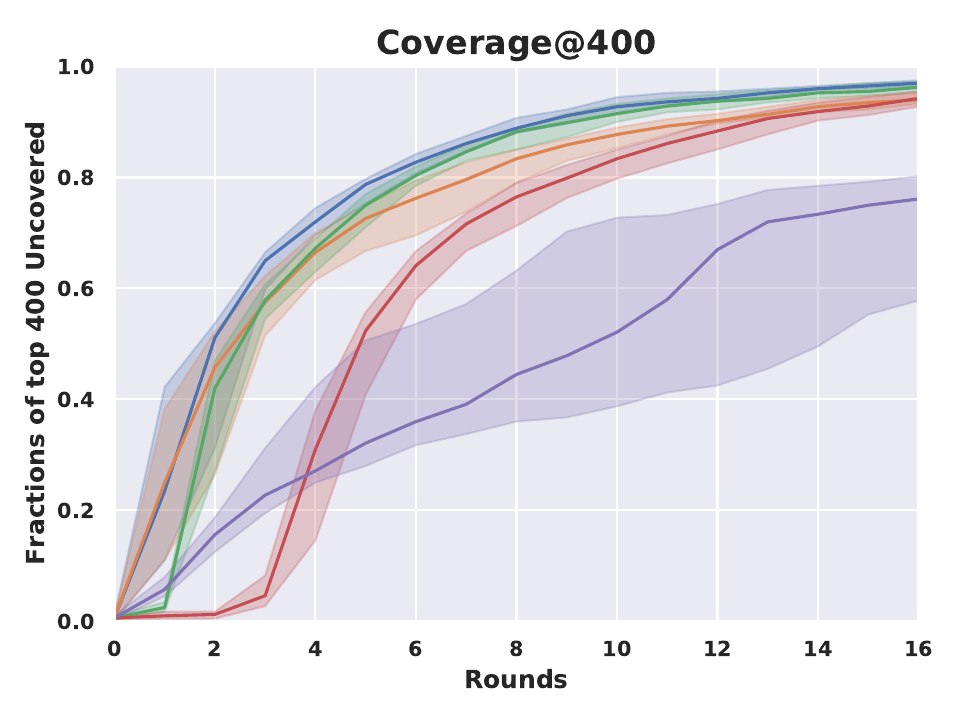}
		\caption{Coverage@400} 
    \end{subfigure}
    \begin{subfigure}{0.9\columnwidth}
        \centering
		\includegraphics[width=\textwidth]{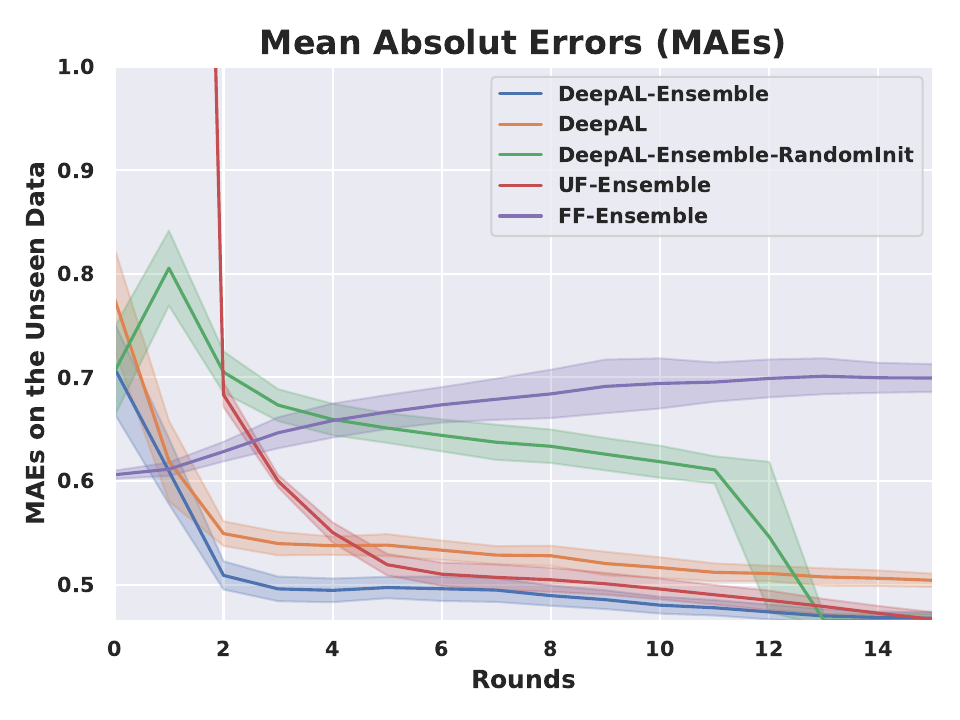}
		\caption{Mean Absolute Errors on Unseen Data}
    \end{subfigure}  
    \caption{Abalation studies to validate our proposed approach, and results are summarized over $20$ replicates. The comparison between the base model and ensemble approaches shows that ensemble method is able to provides a meaningful uncertainty quantification that benefits coverage in the long-term; The comparison between DeepAL-Ensemble versus UF-Ensemble shows that the graph representation of the models from SPOKE offers a more effective representation of the genes for downstream tasks; The comparison between the DeepAL-ensemble and DeepAL-Ensemble with random initialization shows that there is meaningful information in the SPOKE graph that can benefits the search for best genes in the early rounds, but the benefits diminish rapidly as more experimental data is collected; The comparison between DeepAL-Ensemble and FF-Ensemble shows that there is significant benefits to fine-tuning the embeddings as more data is being collected.}
    \label{fig: ablation studies}
\end{figure*}

To demonstrate the significance of the various components of our model, we perform the following ablation studies:
\begin{enumerate}
   \item DeepAL: base model. 
   \item DeepAL-Ensemble: ensemble of base models with $M=20$. 
   \item DeepAL-Ensemble-Randominit: ensemble of the base models but all the weights are initialized randomly instead of performing self-supervised training through negative sampling. 
   \item Ensemble of Models with Fixed Features (FF-Ensemble): this is the case where the embeddings are frozen after initial training on the knowledge graph, and the comparison aims to examine the benefits of fine-tuning the embeddings on the observed viral-load data. 
   \item Ensemble of Models with Unconstrained Features (UF-Ensemble): this is the case where the embeddings are treated as free parameters to be optimized and has been studied extensively to understand behaviors of neural networks \cite{mixon2022neural, zhu2021geometric}. The comparison aims to examine the benefits of incorporating knowledge graph information into the model. 
\end{enumerate}
\noindent In Fig. \ref{fig: ablation studies} we compare our best performing acquisition strategy (optimism with $10\%$-quantile) for all ensemble based approaches listed above and a greedy acquisition strategy for the base model. The comparison between the base model and ensemble approaches show that ensemble methods are able to provide  meaningful uncertainty quantification that benefit coverage in the long-run. The comparison between DeepAL-Ensemble versus UF-Ensemble shows that the graph embeddings of the models extracted from SPOKE offers a more effective representation of the genes for downstream tasks; The comparison between the DeepAL-ensemble and DeepAL-Ensemble with random initialization shows that there is meaningful information in the SPOKE graph that can accelerate the search for best genes in the early rounds when only a small amount of experimental data has been observed; The comparison between DeepAL-Ensemble and FF-Ensemble shows that there is significant benefit in fine-tuning embeddings based on observed experimental data. 

\subsection{Pathway Analysis}
\label{subsec: bio interpretations}

\begin{figure*}[htbp]
    \centering
    \begin{subfigure}{0.9\columnwidth}\hspace*{-1em}
        \centering
        \includegraphics[width=\textwidth]{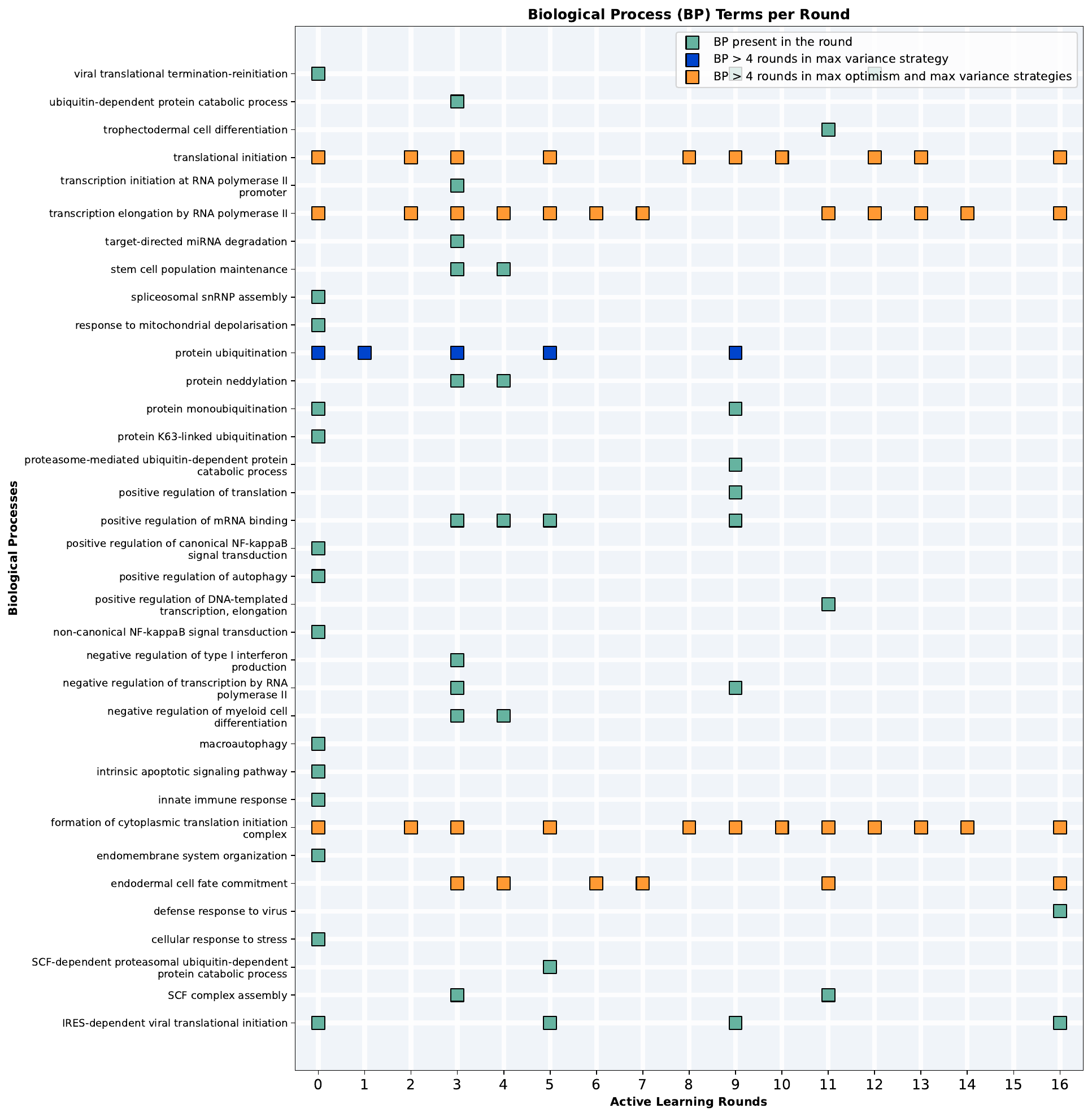}
        \caption{Pathway analysis based on maximum variance.}
    \end{subfigure}
    \begin{subfigure}{0.9\columnwidth}
        \centering
        \includegraphics[width=\textwidth]{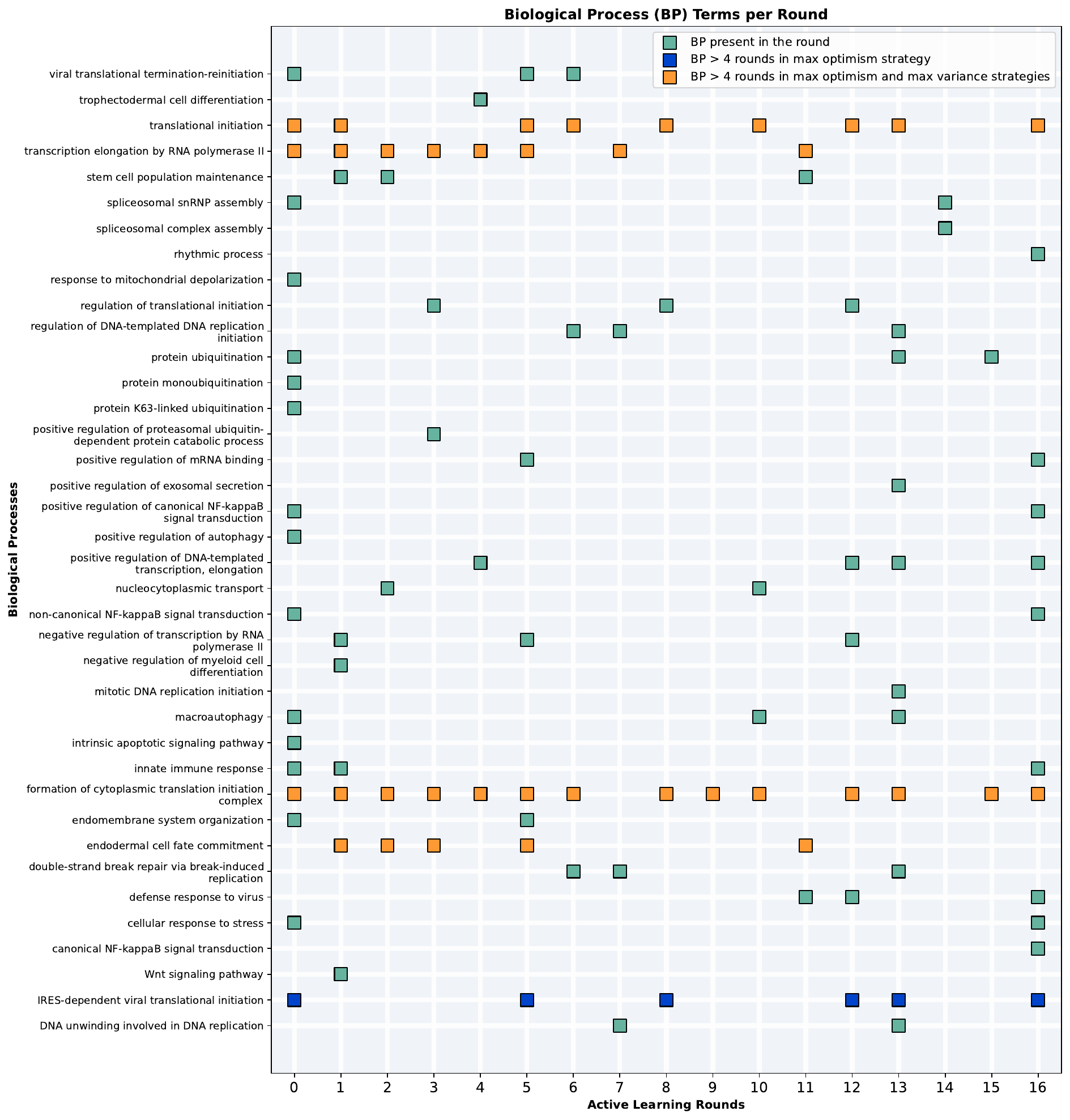}
        \caption{Pathway analysis based on optimism.}
    \end{subfigure}  
    \caption{Pathway enrichment analysis on the gene-pairs selected by DeepAL. The left panel shows the biological process terms selected optimizing by maximum variance, and the right panel shows the terms enriched when optimizing by maximum optimism. The $x$-axis represents the rounds, while the $y$-axis shows the total number of Gene Ontology (GO) biological process terms that are most frequently selected by the algorithm.}
    \label{fig: pathway}
\end{figure*}

To provide a biologically intuitive representation of the gene selection process during active learning, we performed a pathway enrichment analysis, with the results summarized in Fig. \ref{fig: pathway}. This analysis involved identifying the gene pairs most frequently selected in $20$ independent runs and grouping the top pairs for each round. Using GOATOOLS \cite{klopfenstein2018goatools} for gene enrichment, each gene group was tested against all coding genes and overrepresented terms were plotted. We compare the results of optimism (best coverage performance) and maximum variance (best mean absolute error performance). Overall, both strategies retrieve a similar number of biological processes (BPs); however, the maximum variance strategy includes significantly more BPs that appear in only one round (32 compared to 25 for the optimism strategy). This difference can be attributed to the broader focus of the maximum-variance strategy on the entire matrix.

Among the biological processes present in both strategies, those highlighted in orange are of particular importance: almost all are related to translation. These processes are observed across more than four rounds in both strategies, underscoring the ability of both strategies to select genes that directly decrease viral replication by halting translational or pre-translational processes, such as mRNA elongation by RNA polymerase II. This is significant in HIV infection as Tat is known to enhance recruitment and activity of RNA polymerase II \cite{reeder2015hiv}.  Among the 30\% of BPs that differ between the strategies, DNA-related processes are explored more extensively by the maximum variance strategy, which identifies three times as many DNA-related BPs compared to the optimism strategy. For RNA- and transcription-related processes, the maximum-variance strategy notably identifies the "positive regulation of mRNA binding" BP in nine rounds. This BP is associated with processes that activate or increase the frequency, rate, or extent of mRNA binding. In contrast, the optimism strategy identifies the "positive regulation of DNA-templated transcription, elongation" BP over five rounds, a process that enhances the frequency of transcription elongation. Both BPs include genes whose silencing may decrease viral replication.

In conclusion, both strategies successfully identify genes that can decrease viral replication, although with differences in focus and approach.


\section{Conclusion}
\label{sec: conclusion}

In this work, an integrated Deep Active Learning (DeepAL) framework is proposed to incorporate existing heterogeneous knowledge graph information for efficient double-gene knockdown experiment design for host-targeted therapeutics. This unified framework utilizes the expressiveness of graph representation learning, the uncertainty quantification of the ensemble method, and the sample efficiency of the active learning loop to show promising results on double-gene knockdown experimental data of appreciable scale. In addition, the framework can provide meaningful representations of genes, leading to interpretable results on the set of gene pairs recommended by the framework. 

There are several directions for future work. One direction is to integrate pre-trained large language models into the framework for initial recommendation steps, since these pre-trained models have demonstrated the ability to provide valuable information for a warm start; the second direction is to incorporate additional information from the knowledge graph into the model (e.g., node features) to better differentiate the gene nodes; a third direction is to develop an efficient Graph Learning inference method through neighbor subsampling which enables us to utilize the full knowledge graph into our model instead of a subgraph.    

\section{Competing interests}
No competing interest is declared.

\section{Author contributions statement}

H.Z., M.S., J.C.P and M.L. contributed to the implementation of the framework, H.Z., P.R., and B.S. contributed to the development of the methodology of the active learning framework. B.P. and S.B. contributed to the biological interpretation of the findings. J.D. contributed to the process of obtaining and preprocessing of the HIV dataset. All authors collaboratively wrote and reviewed the manuscript.

\section{Acknowledgments}
This work was performed under the auspices of the U.S. Department of Energy by Lawrence Livermore National Laboratory under Contract DE-AC52-07NA27344(LLNL-JRNL-872100). Funding was provided by the Lawrence Livermore National Laboratory Directed Research and Development project 23-SI-005.

\bibliographystyle{plain}
\bibliography{main}

\begin{biography}{{\color{black!20}\rule{77pt}{77pt}}}
{\author{Haonan Zhu}
received the Ph.D. degree in electrical and computer engineering from the University of Michigan in 2023. He is currently a Postdoctoral
Researcher with the Lawrence Livermore National Laboratory.} 
\end{biography}

\begin{biography}{{\color{black!20}\rule{77pt}{77pt}}}
{\author{Mary Silva} received an M.S. in Statistics from University of California, Santa Cruz. She is currently a Data Scientist in the Biomolecular Design and Development Group at Lawrence Livermore National Laboratory.} 
\end{biography}

\begin{biography}{{\color{black!20}\rule{77pt}{77pt}}}
{\author{Jose Cadena} received the Ph.D. degree in computer science from the Department of Computer Science at Virginia Tech. He is currently a Research Staff Member with the Computational Engineering Division at Lawrence Livermore National Laboratory.} 
\end{biography}

\begin{biography}{{\color{black!20}\rule{77pt}{77pt}}}
{\author{Braden Soper} received the Ph.D. degree in Applied Mathematics and Statistics from the University of California, Santa Cruz. He is currently Data Scientist at Lawrence Livermore National Laboratory.} 
\end{biography}

\begin{biography}{{\color{black!20}\rule{77pt}{77pt}}}
{\author{Michał Lisicki} is completing his Ph.D. at the University of Guelph and the Vector Institute in Ontario, Canada, focusing on sequential decision making. He contributed to the present work during his internship at Lawrence Livermore National Laboratory.} 
\end{biography}

\begin{biography}{{\color{black!20}\rule{77pt}{77pt}}}
{\author{Braian Peetoom} earned an M.D. from Universidad Nacional de La Plata in 2022. He is currently a Postdoctoral Researcher at Baranzini Lab in UCSF.} 
\end{biography}

\begin{biography}{{\color{black!20}\rule{77pt}{77pt}}}
{\author{Sergio E Baranzini} received his Ph.D. degree in molecular human genetics from the University of Buenos Aires in 1997. He is currently a distinguished Professor of Neurology at the University of California San Francisco.} 
\end{biography}

\begin{biography}{{\color{black!20}\rule{77pt}{77pt}}}
{\author{Priyadip Ray} received the Ph.D. degree in electrical
engineering from Syracuse University in 2009. He is
currently a Staff Scientist with the Machine Learning Group at Lawrence Livermore National Laboratory.}
\end{biography}

\begin{biography}{{\color{black!20}\rule{77pt}{77pt}}}
{\author{Jeff Drocco} completed his Ph.D. in physics at Princeton University in 2011. He is the Deputy Group Leader for Genomics in the Biosciences and Biotechnology Division at Lawrence Livermore National Laboratory.}
\end{biography}

\end{document}


\journaltitle{ISMB 2025 Proceeding}
\DOI{DOI}
\copyrightyear{2025}
\pubyear{2015}
\access{}
\appnotes{Paper}

\firstpage{1}

\title[Deep Active Learning for Host Targeted Therapeutics (Supplementary)]{Supplementary: Deep Active Learning based Experimental Design to Uncover Synergistic Genetic Interactions for Host Targeted Therapeutics}

\author[1,$\ast$]{Haonan Zhu}
\author[1]{Mary Silva}
\author[1]{Jose Cadena} 
\author[1]{Braden Soper}
\author[2,3]{Michał Lisicki}
\author[4]{Braian Peetoom}
\author[4]{Sergio E. Baranzini} 
\author[1]{Shankar Sundaram}
\author[1]{Priyadip Ray}
\author[1]{Jeff Drocco}
\authormark{Haonan Zhu et al.}

\address[1]{\orgname{Lawrence Livermore National Laboratory}, \orgaddress{\street{7000 East Ave}, \postcode{94550}, \state{CA}, \country{USA}}}
\address[2]{ \orgname{University of Guelph}, \orgaddress{\street{50 Stone Rd E}, \postcode{N1G 2W1}, \state{ON}, \country{Canada}}}
\address[3]{  \orgname{Vector Institute}, \orgaddress{\street{108 College St W1140}, \postcode{M5G 0C6}, \state{ON}, \country{Canada}}}
\address[4]{  \orgname{University of California San Francisco}, \orgaddress{\street{675 Nelson Rising Lane}, \postcode{94158}, \state{CA}, \country{USA}}}

\corresp[$\ast$]{Corresponding author. \href{email:zhu18@llnl.gov}{zhu18@llnl.gov}}

\begin{abstract}
\mbox{}
\end{abstract}

\maketitle

\section{Additional Experimental Results}
\label{sec: additional experimental results}

In this section, we provide additional experimental results on the HIV dataset \cite{gordon2020quantitative}. 

\subsection{Results for Batch Size=200}
\label{subsec: batch size 200}

In this subsection, we provide additional results where we run all algorithms for $33$ rounds (including the initial round of random selection), and at each round each algorithm selects $200$ gene pairs for observation,

\begin{figure}[h]
    \centering
    \begin{subfigure}{0.45\columnwidth}\hspace*{-1em}
        \centering
		\includegraphics[width=\textwidth]{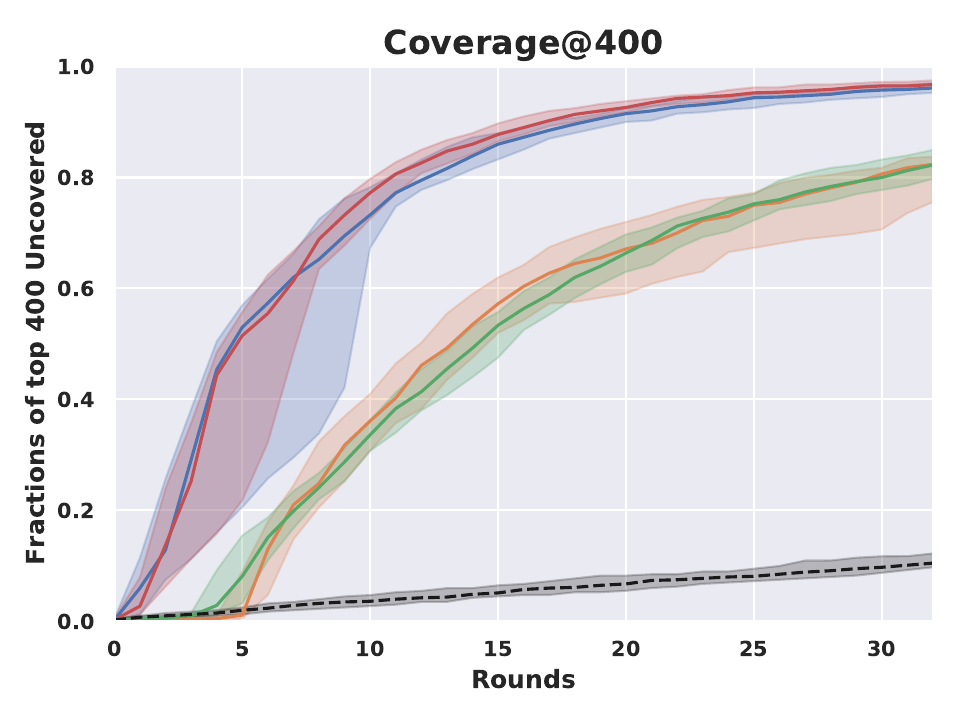}
		\caption{Coverage@400} 
    \end{subfigure}
    \begin{subfigure}{0.45\columnwidth}
        \centering
		\includegraphics[width=\textwidth]{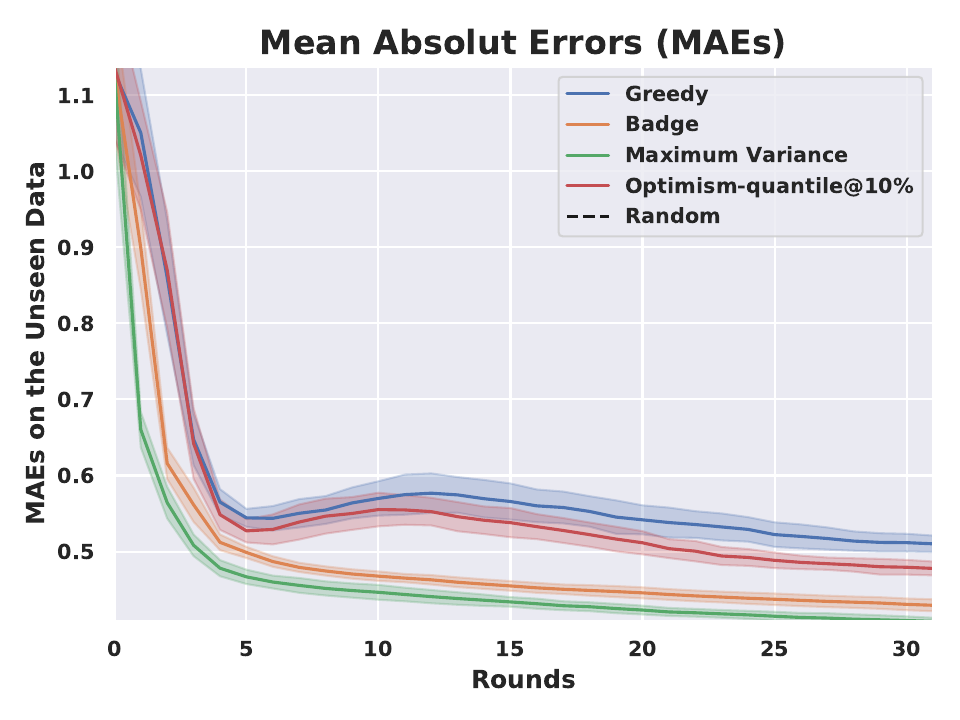}
		\caption{Mean Absolute Errors on Unseen Data}
    \end{subfigure}  
    \caption{Comparison among the different acquisition strategies, and results are summarized over $20$ replicates. Optimism with $10\%$ quantiles achieves the best performance in terms of coverage at terminal phase. Maximum variance strategy performs the best in learning a generalizable model that predicts viral-replicates on unseen data due to an emphasis on exploration.}
    \label{fig: acquisition comparisons batsize 200}
\end{figure}

\begin{figure}[h]
    \centering
    \begin{subfigure}{0.45\columnwidth}\hspace*{-1em}
        \centering
		\includegraphics[width=\textwidth]{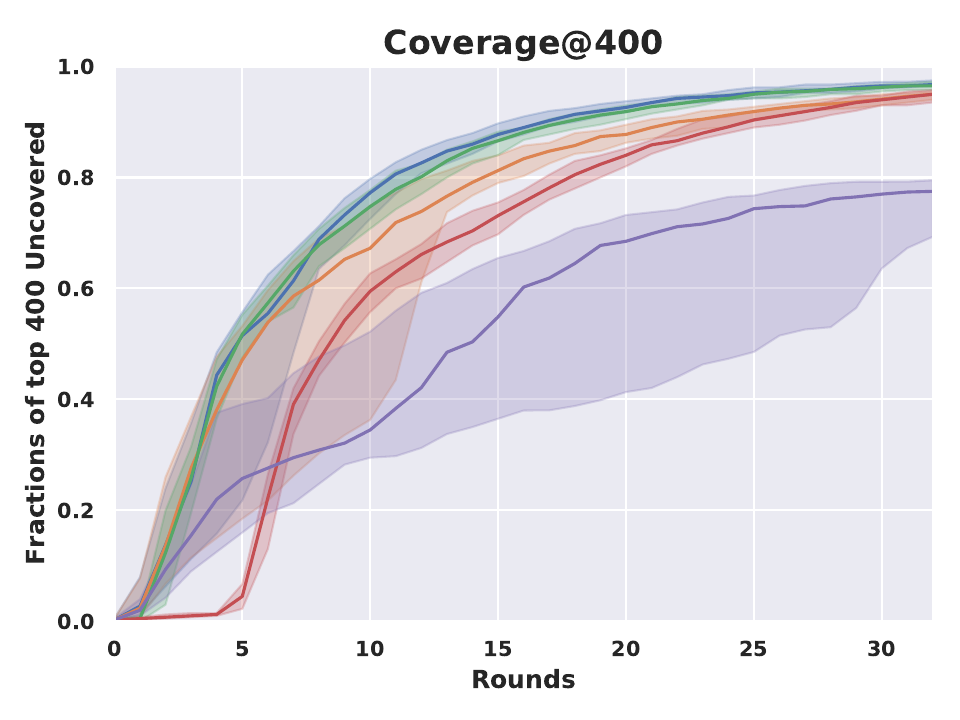}
		\caption{Coverage@400} 
    \end{subfigure}
    \begin{subfigure}{0.45\columnwidth}
        \centering
		\includegraphics[width=\textwidth]{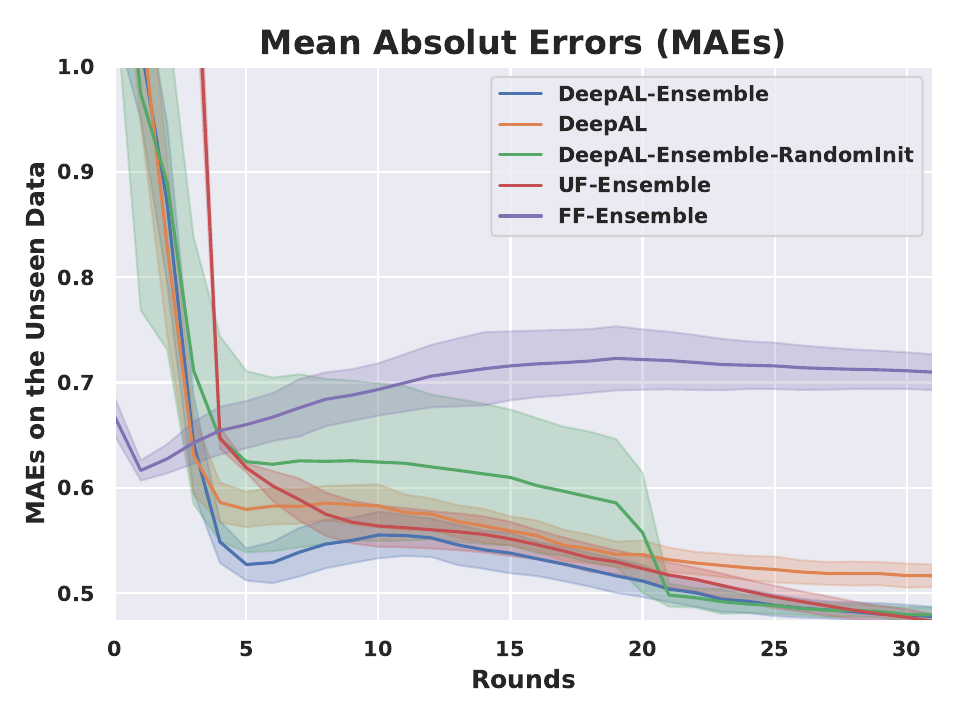}
		\caption{Mean Absolute Errors on Unseen Data}
    \end{subfigure}  
    \caption{Abalation studies to validate our proposed approach, and results are summarized over $20$ replicates.}
    \label{fig: ablation studies batsize 200}
\end{figure}

\subsection{Results for Batch Size=800}
\label{subsec: batch size 800}

In this subsection, we provide additional results where we run all algorithms for $9$ rounds (including the initial round of random selection), and at each round each algorithm selects $800$ gene pairs for observation,

\begin{figure}[h]
    \centering
    \begin{subfigure}{0.45\columnwidth}\hspace*{-1em}
        \centering
		\includegraphics[width=\textwidth]{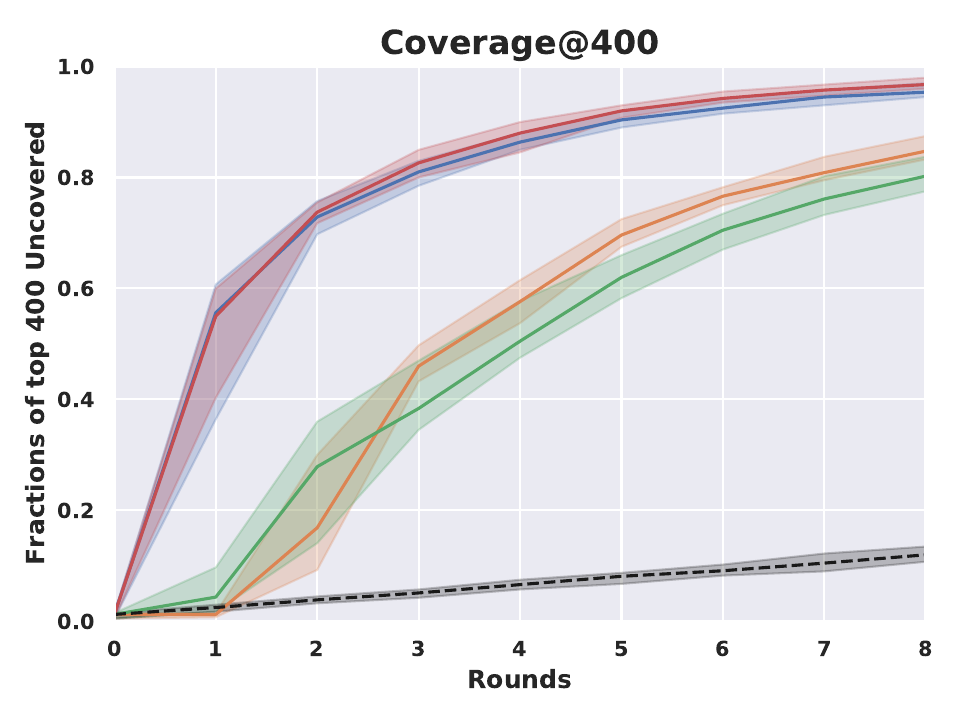}
		\caption{Coverage@400} 
    \end{subfigure}
    \begin{subfigure}{0.45\columnwidth}
        \centering
		\includegraphics[width=\textwidth]{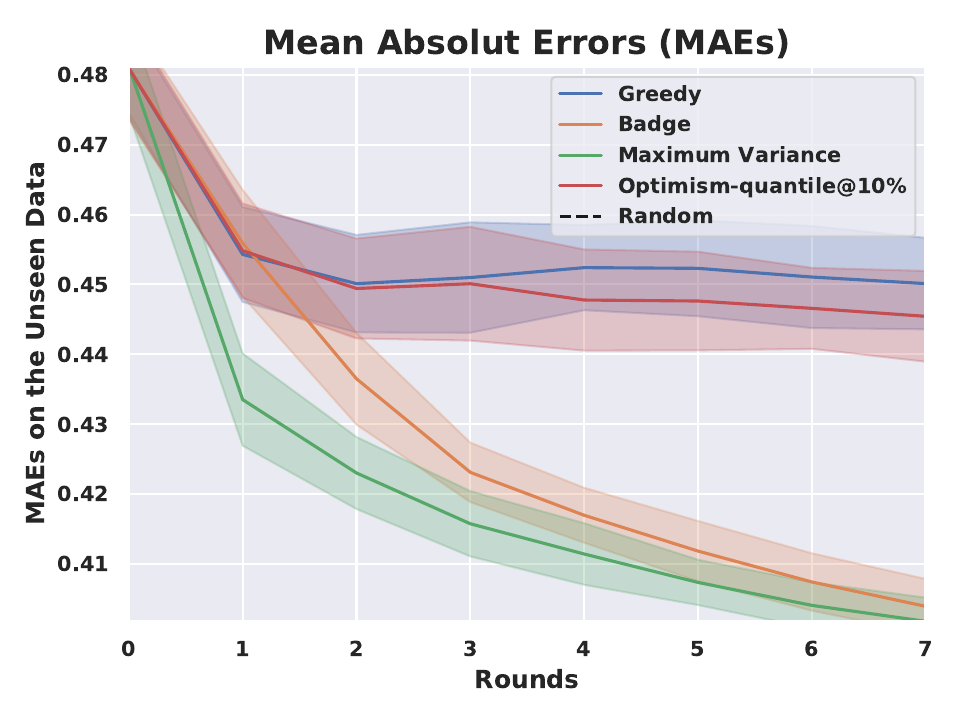}
		\caption{Mean Absolute Errors on Unseen Data}
    \end{subfigure}  
    \caption{Comparison among the different acquisition strategies, and results are summarized over $20$ replicates. Optimism with $10\%$ quantiles achieves the best performance in terms of coverage at terminal phase. Maximum variance strategy performs the best in learning a generalizable model that predicts viral-replicates on unseen data due to an emphasis on exploration.}
    \label{fig: acquisition comparisons batsize 800}
\end{figure}

\begin{figure}[h]
    \centering
    \begin{subfigure}{0.45\columnwidth}\hspace*{-1em}
        \centering
		\includegraphics[width=\textwidth]{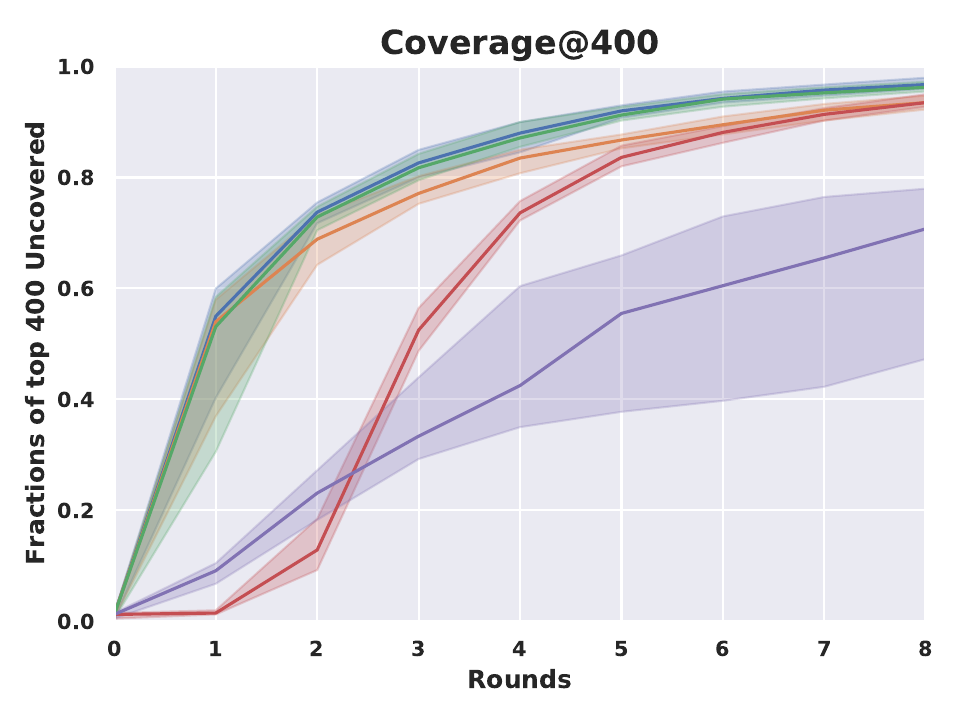}
		\caption{Coverage@400} 
    \end{subfigure}
    \begin{subfigure}{0.45\columnwidth}
        \centering
		\includegraphics[width=\textwidth]{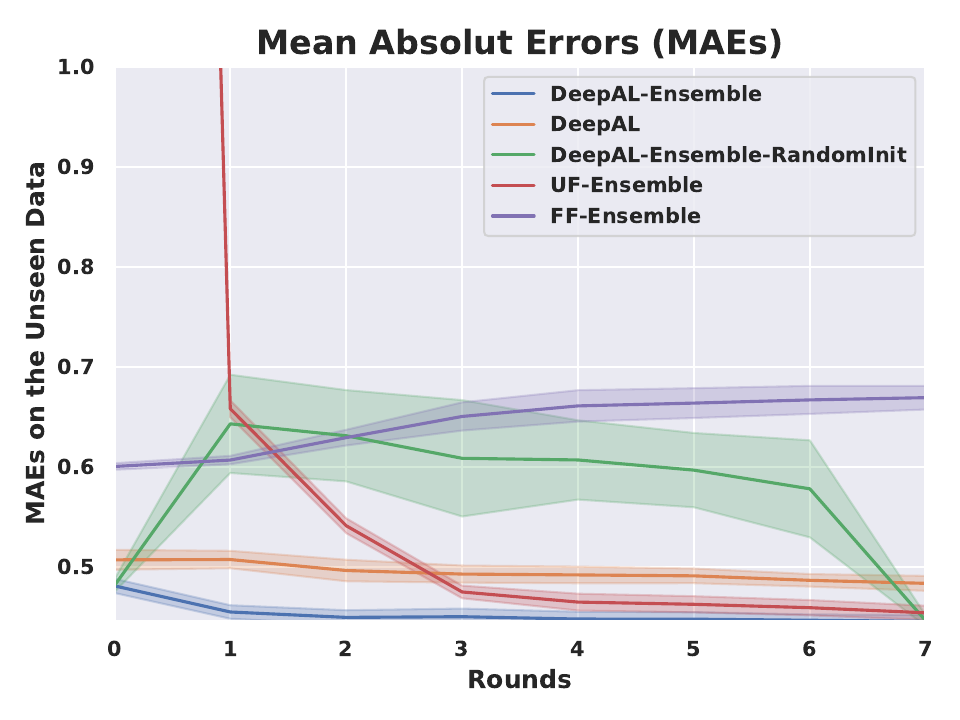}
		\caption{Mean Absolute Errors on Unseen Data}
    \end{subfigure}  
    \caption{Abalation studies to validate our proposed approach, and results are summarized over $20$ replicates.}
    \label{fig: ablation studies batsize 800}
\end{figure}

\bibliographystyle{plain}
\bibliography{main}